\documentclass[letterpaper, 10 pt, conference]{ieeeconf}
\usepackage{amssymb}
\usepackage{arydshln}
\usepackage{colortbl}
\usepackage{dashrule}
\usepackage{amsmath}
\usepackage{hyperref}
\usepackage{graphicx}
\usepackage{booktabs}
\usepackage{multirow}
\usepackage{subcaption}
\usepackage{hyperref}

\usepackage{siunitx}
\usepackage{caption}

\usepackage{array}
\usepackage{etoolbox}
\usepackage{comment}
\usepackage{arydshln}
\makeatletter
\patchcmd{\@makecaption}
  {\scshape}
  {}
  {}
  {}
\makeatletter
\patchcmd{\@makecaption}
  {\\}
  {.\ }
  {}
  {}
\makeatother

\usepackage{colortbl}

\usepackage[dvipsnames]{xcolor}

\DeclareMathOperator*{\argmax}{arg\,max}

\DeclareMathOperator{\Tr}{Tr}

\newcommand{\R}{\mathbb{R}}

\newcommand{\X}{\mathcal{X}}
\newcommand{\Y}{\mathcal{Y}}

\IEEEoverridecommandlockouts
\title{\LARGE \bf
OffRIPP: Offline RL-based Informative Path Planning
}

\author{Srikar Babu Gadipudi$^{1}$, Srujan Deolasee$^{2}$, Siva Kailas$^{3}$, Wenhao Luo$^{4}$, Katia Sycara$^{2}$, Woojun Kim$^{2}$
\thanks{$^{1}$Srikar Babu Gadipudi is with the Department of Electrical Engineering, Indian Institute of Technology Madras. ${\tt\small ee21b138@smail.iitm.ac.in}$
        }
\thanks{$^{2}$Srujan Deolasee, Katia Sycara, and Woojun Kim are with the Robotics Institute at Carnegie Mellon University. ${\tt\small \{sdeolase, sycara, woojunk \}@andrew.cmu.edu}$
        }
\thanks{$^{3}$Siva Kailas is with the School of Interactive Computing at Georgia Institute of Technology ${\tt\small \{skailas3\}@gatech.edu}$}
\thanks{$^{4}$Wenhao Luo is with the Department of Computer Science, University of North Carolina at Charlotte
        ${\tt\small \{wenhao.luo\}@uncc.edu}$}
}

\begin{document}

\maketitle
\thispagestyle{empty}
\pagestyle{empty}

\begin{abstract}

Informative path planning (IPP) is a crucial task in robotics, where agents must design paths to gather valuable information about a target environment while adhering to resource constraints. Reinforcement learning (RL) has been shown to be effective for IPP, however, it requires environment interactions, which are risky and expensive in practice. To address this problem, we propose an offline RL-based IPP framework that optimizes information gain without requiring real-time interaction during training, offering safety and cost-efficiency by avoiding interaction, as well as superior performance and fast computation during execution---key advantages of RL. Our framework leverages batch-constrained reinforcement learning to mitigate extrapolation errors, enabling the agent to learn from pre-collected datasets generated by arbitrary algorithms. We validate the framework through extensive simulations and real-world experiments. The numerical results show that our framework outperforms the baselines, demonstrating the effectiveness of the proposed approach. 

\end{abstract}

\begin{keywords}
    Informative Path Planning; Offline Reinforcement Learning
\end{keywords}

\section{INTRODUCTION}
Informative path planning (IPP) is a critical problem in robotics and autonomous systems, where the goal is to design a path that enables an agent to acquire valuable information about a quantity of interest within a given environment while adhering to resource constraints, such as a robot's battery life. IPP has numerous applications, including environmental monitoring~\cite{hitz2017adaptive}, search and rescue operations~\cite{meera2019obstacle}, and precision agriculture~\cite{popovic2020informative}. The challenge is to ensure that the agent accurately approximates the true interest map.

Traditional methods for IPP, \cite{karaman2011sampling, arora2017randomized}, rely on sampling techniques that, while effective, can be computationally intensive and time-consuming. They typically require significant computational resources and often struggle to scale in large, complex environments. Consequently, there is growing interest in leveraging reinforcement learning (RL) as a potential solution to the IPP problem \cite{cao2023catnipp, ruckin2022adaptive, vashisth2024deep}. RL offers a promising avenue for IPP by enabling agents to learn optimal policies through interactions with the environment. However, conventional RL approaches require extensive real-time interactions, making training costly and potentially hazardous, especially in safety-critical environments. To address this, offline RL, which trains an agent using only pre-collected datasets without real-time interactions, has emerged as a viable alternative~\cite{kumar2020conservative}.

\begin{figure}
    \centering
    \includegraphics[width=\linewidth]{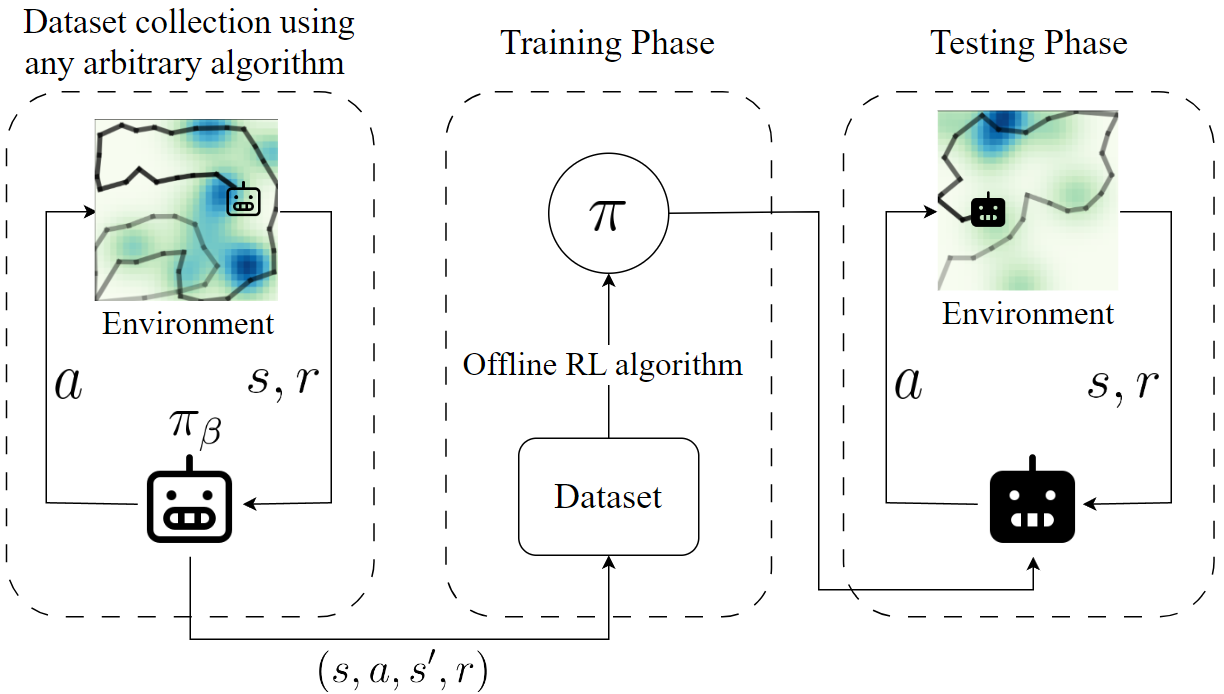}
    \caption{Overview of the flow for the proposed framework: training an RL policy using a dataset generated by arbitrary algorithms, followed by deployment in the test environment.}
    \label{fig:flow}
    \vspace{-2ex}
\end{figure}

In this paper, we propose an \textbf{Off}line \textbf{R}L-based \textbf{IPP} framework, named \texttt{OffRIPP}, that plans paths to maximize information gain without real-time interactions with the environment during training. We train an RL agent to maximize information gain within a constrained budget, using datasets generated by arbitrary IPP algorithms, including traditional sampling-based approaches. Specifically, we leverage batch-constrained reinforcement learning~\cite{fujimoto2019off} for IPP to mitigate extrapolation errors---these occur due to inaccurate value estimation caused by the mismatch between the state-action distribution in the dataset and the current policy. After training, the RL policy can be deployed directly in the test environment. The overall flow is illustrated in Fig. \ref{fig:flow}. Our framework provides an efficient and safe solution, combining the advantages of RL---fast planning time during execution and superior performance---with the benefits of traditional approaches that require no interaction with the environment.  

We validate our approach through extensive simulations and real-world experiments, demonstrating the advantages of our offline RL framework in improving performance and reducing operational costs compared to several baseline methods. We evaluate the proposed framework on top of two existing RL-based IPP algorithms in two different environments: a 2D light-intensity task and a 3D fruit identification task. The numerical results show that our framework outperforms the baselines, including traditional approaches, online RL-based algorithms trained in an offline manner, and behavior cloning, demonstrating the effectiveness of the proposed framework. 

The main contributions of this work are summarized as follows: (i) To the best of our knowledge, this is the first work to leverage offline RL for IPP, training solely on given datasets. (ii) Our approach can be integrated with any existing online RL-based IPP algorithm; we demonstrate this by using both CAtNIPP~\cite{cao2023catnipp} and another RL-based IPP~\cite{vashisth2024deep} as subroutines. (iii) We demonstrate the effectiveness of our approach in terms of solution quality and planning time, and provide an analysis of dataset quality.

\section{Background and Related works} \label{sec:related_works}

\subsection{Offline Reinforcement Learning}\label{sec:offlineRL}

\textbf{RL} ~trains an agent through interaction with an environment, and the procedure is formulated as a Markov Decision Process, defined by the tuple $<\mathcal{S}, \mathcal{A}, \mathcal{P}, r, \gamma>$, where $\mathcal{S}$ is the state space, $\mathcal{A}$ is the action space, $\mathcal{P}$ is the transition probability function, $r$ is the reward function, and $\gamma$ is the discount factor. Specifically, at each time step $t$, the agent's policy $\pi$ selects an action $a_t \in \mathcal{A}$ based on the current state $s_t \in \mathcal{S}$. The environment responds by providing a reward $r_t = r(s_t,a_t)$ and transitioning to the next state $s_{t+1} \sim \mathcal{P}(s_{t+1} | s_t,a_t)$. By repeating this procedure in an online learning fashion, the policy is trained to maximize the expected return, $\mathbb{E}[\sum_{t=0}^{\infty} \gamma^t r_t]$. Here, to evaluate and improve the policy, most RL algorithms estimate the value function under the current policy, such as $Q^{\pi}(s,a)=\mathbb{E}_{\pi}[\sum_{t=0}^{\infty} \gamma^t r_t \mid s, a]$, which is the expected return over the trajectory that follows policy $\pi$ after executing action $a$ in state $s$. 
One way to estimate this is temporal-difference (TD) learning~\cite{tesauro1995temporal}, which can be written as
\begin{align}\label{eq:td}
    Q^{\pi}(s,a) \leftarrow \mathbb{E}_{s'}\left[ r+\gamma Q^{\pi}(s', \pi(s'))
 \right],
\end{align}
where $s'$ is the next state. The policy is trained to maximize the estimated value function. Thus, accurate estimation of the value function is crucial, as it directly affects policy learning.

\textbf{Offline RL} ~trains an agent on a fixed dataset previously collected from arbitrary policies, without any interaction with the environment during training~\cite{kim2024value, kumar2020conservative, kim2024decision}. By avoiding interaction, which is often risky and expensive in real-world applications, offline RL enables the development of a decision-making agent that maximizes the expected sum of rewards. A significant challenge in offline RL is extrapolation error caused by the mismatch between the dataset and the actual state-action distribution of the current policy. This leads to inaccurate value estimates~\cite{fujimoto2019off}. For example, in TD learning, if a state-action pair $(s', \pi(s'))$ is not included in the dataset, the estimate of the value function for this state-action pair can be highly inaccurate. In other words, the expectation in Eq. \ref{eq:td}
has a large approximation error, which accumulates in the estimate of $Q^{\pi}(s,a)$ in Eq. \ref{eq:td}. Thus, a greater mismatch between the dataset and the state-action distribution of the current policy increases the likelihood of approximation error. 

To handle extrapolation error, most offline RL algorithms make the learning policy similar to the behavior policy, which was used to generate the dataset~\cite{fujimoto2019off, kostrikov2021offline, kumar2020conservative}, in order to reduce the mismatch. For instance, Conservative Q-learning penalizes Q-values for actions not seen in the dataset, thereby discouraging the policy from selecting out-of-distribution actions~\cite{kumar2020conservative}. TD3+BC~\cite{fujimoto2021minimalist} integrates TD3~\cite{fujimoto2018addressing} with behavior cloning, combining exploration and exploitation while staying close to the dataset's actions. Batch-constrained Q-learning constrains the policy to select actions observed in the dataset~\cite{fujimoto2019off}.

\subsection{Informative Path Planning}\label{subsec:ipp}

The goal of IPP is to find a trajectory $\psi^*$ that maximizes the information gain within the given budget~\cite{ruckin2023informative, kailas2023multi, popovic2024learning}, and the objective function is written as
\begin{equation}
    \psi^* = \argmax_{\psi} I(\psi), \text{   s.t. } C(\psi) \le B, 
\end{equation}
where $I: \psi \rightarrow \R^+$ and $C: \psi \rightarrow \R^+$ represent the information gain and cost along trajectory $\psi$. $B \in \R^+$ is the limit of the budget. The type of information gain depends on the domain, e.g., fire intensity in the wildfire domain. To find a solution for IPP, traditional IPP algorithms utilize sampling methods~\cite{karaman2011sampling, jones2013receding, yoo2016experimental, hitz2017adaptive}. However, they require significant computational resources for planning, which limits their practicality in real-world applications.

\textbf{RL-based IPP:} RL has been utilized to learn an IPP solver to address the aforementioned problem of traditional IPP solvers and further enhance performance~\cite{wei2020informative,  cao2023catnipp, vashisth2024deep}. RL-based IPP algorithms consist of three components: constructing a representation of the entire search map, modeling environmental phenomena, and training an RL agent to select a path that maximizes information gain.  As an example, CAtNIPP~\cite{cao2023catnipp} constructs a representation to cover a continuous 2D search domain using a probabilistic roadmap (PRM)~\cite{geraerts2004comparative}, which reduces the complexity of the search space. At the start of each episode, the PRM generates a graph $G = (V, E)$ with nodes $V$ and edges $E$, where each node is connected to $k$ neighbors, and the agent is initialized at a random node. At each time step, the agent moves to one of its neighboring nodes and observes the environmental phenomenon at its new location. To model this phenomenon across the search space, Gaussian Process (GP) regression, a non-parametric Bayesian method that uses statistical inference to capture relationships between data points, is used~\cite{seeger2004gaussian, wang2023intuitive}. Specifically, the mean and covariance of a test location $X^*$, given the observed locations $X$, are inferred as: $\mu = K(\X^*,\X)[K(\X,\X) + \sigma_n^2I]^{-1}(\Y - \mu(\X)), P = K(\X^*, \X^*) - K(\X^*, \X)[K(\X,\X) + \sigma_n^2I]^{-1}K(\X,\X)^T$, where $K(\cdot, \cdot)$ is the kernel function, $\sigma_n^2$ is a hyperparameter of the GP. At each time step, the agent predicts the environmental phenomenon based on the GP and uses this prediction as input to the RL policy. After observing the environmental phenomenon at the new location, the GP is updated based on the new observation value. The output of the GP is incorporated into the graph, referred to as the GP-augmented graph. The RL policy uses the agent's current location, budget, trajectory history, and the GP-augmented graph, as inputs. The RL agent then selects one of the neighboring nodes. To train the agent to maximize the expected reduction in GP uncertainty, the reward function is designed as the normalized information gain, $r_t = (\Tr(P^{t-1}) - \Tr(P^t))/\Tr(P^{t-1})$, with a negative reward term, $r_d = -\alpha \cdot \Tr(P^d)$, added at the end of each episode. CAtNIPP uses PPO~\cite{schulman2017proximal} for training. 

As another example, \cite{vashisth2024deep} proposes an RL-based IPP algorithm for 3D environments with unknown obstacles. To avoid collisions, they introduce a dynamically constructed graph representing the robot’s local region, which is used as input to the RL agent. Additionally, they propose a new reward function to balance exploration and exploitation. Apart from these components, this 3D RL-based IPP algorithm follows the CAtNIPP framework.

These RL-based IPP methods has been shown to be effective in terms of both solution quality and planning time in test time. However, they require interaction with the environment to collect data which is expensive, risky, and time-consuming. Thus, we introduce offline RL to the IPP problem. To the best of our knowledge, this work is the first to develop an offline RL-based IPP solver.

\begin{figure*}
    \centering
    \includegraphics[width=0.9\textwidth]{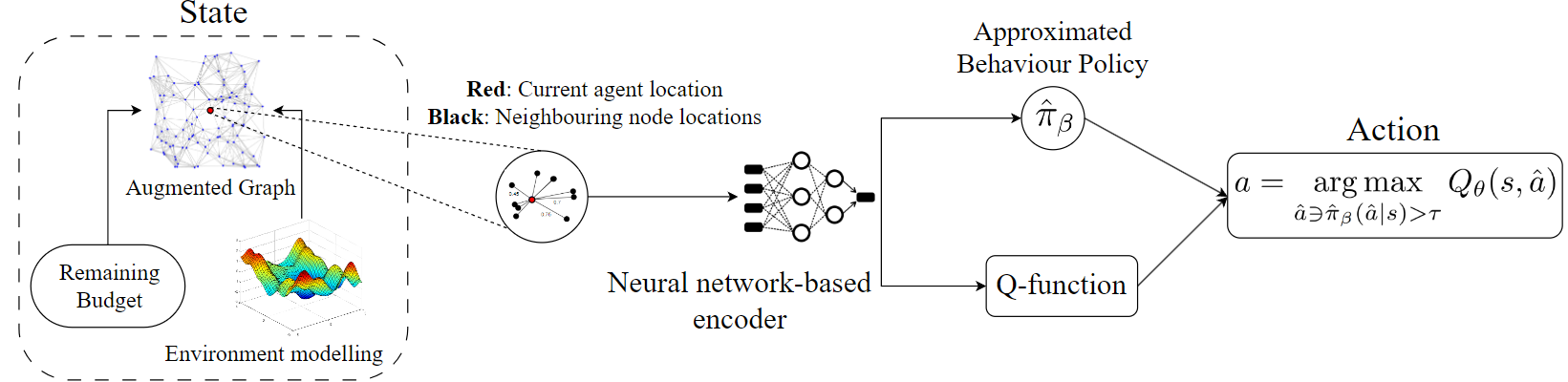}
    \caption{The architecture of \texttt{OffRIPP}: A graph augmented by environment modeling (the output of GP) and the remaining budget are used as input. The approximated behavior policy and Q-function are used to determine an action.}
    \label{fig:Sys_model}
\end{figure*}

\section{Our approach} \label{sec:offline_rl_based_ipp}

We propose an offline RL-based Informative Path Planning algorithm, \texttt{OffRIPP}, that trains an IPP solver using existing datasets without requiring interaction with the environment. By avoiding costly and risky real-time interactions, \texttt{OffRIPP} is applicable to real-world scenarios. In addition, \texttt{OffRIPP} can be combined with any arbitrary RL-based IPP algorithm, including CAtNIPP~\cite{cao2023catnipp} and 3D RL-based IPP~\cite{vashisth2024deep}.

\subsection{Problem Formulation}

\textbf{Dataset.} The dataset $\mathcal{D}$ is assumed to be generated by arbitrary algorithms and consists of $D$ episodes, with each episode containing sequences of states, actions, and rewards. Here, the states include a graph where each node contains a position and the agent's prediction of the environment phenomenon, as well as the planning state, which includes the current agent position and the remaining budget. The actions represent the next node the agent moves to, and the rewards correspond to the GP uncertainty. 

By utilizing the dataset, our goal is to train an RL agent that plans a path maximizing the information gain under the budget constraints. Note that given that the dataset includes PRMs and predictions of environmental phenomena through GP regression, our approach does not require generating these components during training but instead leverages the given dataset. Upon completion of the RL agent's training, we deploy it in the test environment for evaluation. 

\subsection{Offline RL-based Informative path planning}

It is known that naive training of RL algorithms, such as PPO~\cite{schulman2017proximal}, on the dataset can lead to inaccuracies in value estimation, resulting in unstable learning, as discussed in Sec. \ref{sec:offlineRL}. We observed that CAtNIPP, which relies on PPO, fails to train on the dataset, as we will discuss in Sec. \ref{subsec:results}. In order to address this instability, we leverage a key concept from batch-constrained Q-learning (BCQ)~\cite{fujimoto2019off}, which restricts the actions selected by the policy during training to a subset of the given dataset. To achieve this, a model is introduced to generate actions similar to those in the batch, which are then used to build a policy with the Q-function. Consequently, \texttt{OffRIPP} consists of two components: an RL agent and a behavior policy approximator. To boost sample efficiency, both modules share the entire neural network-based function approximators, except for the final layer.

\subsubsection{Behavior Policy Approximation} We build an approximation of the behavior policy, $\hat{\pi}_{\theta_{\beta}}$, to ensure that the learning policy avoids selecting actions that are not supported by the dataset. Since we only have access to the dataset generated by the behavior policy, and not the behavior policy itself, we utilize imitation learning, which minimizes the negative log-likelihood function. The loss function for the behavior policy approximator, parameterized by $\theta_{\beta}$, is written as:
\begin{equation}
    \mathcal{L}({\theta_\beta}) = - \mathbb{E}_{(s,a)\sim \mathcal{D}}\left[a \log \hat{\pi}_{\theta_{\beta}} (a | s)\right],
\end{equation}
where $(s,a)$ is the state-action pair in the dataset. We pretrain the behavior policy approximator before policy learning.

\subsubsection{Policy Learning} We build a policy based on the Q-function, $Q_{\theta_Q}(s, a)$, and train it using the TD update. Here, in contrast to Q-learning, which employs a greedy policy, represented as $\pi(s) = \argmax Q(s, a)$, following BCQ, we use the behavior policy approximator to construct a policy based on the Q-function as follows:
\begin{equation}\label{eq:ourpolicy}
    \pi(s) = \argmax_{\hat{a} \ni \hat{\pi}_{\beta} (\hat{a} | s) > \tau} Q_\theta(s, \hat{a}),
\end{equation}
where $\tau$ is the threshold that defines how much the learning policy deviates from the behavior policy. In other words, we restrict the policy to generate actions that the behavior policy is likely to generate with a probability above the threshold $\tau$. Note that $\tau = 0$ corresponds to fully imitating the behavior policy, while $\tau = 1$ corresponds to following the greedy policy. The rationale behind this is to avoid using unseen state-action pairs, which can lead to extrapolation errors. Based on this policy, we train the Q-function using TD update and the corresponding loss function is written as
\begin{align}
    \mathcal{L}(\theta_{Q}) = \mathbb{E}_{(s,a,r,s')\sim \mathcal{D}} \Big[ \Big(r + \gamma & \max_{\hat{a} \ni \hat{\pi}_{\beta} (\hat{a} | s') > \tau} Q_{\bar{\theta}_Q} (s', \hat{a}) \nonumber \\ &~~~~~~~ - Q_{\theta_Q} (s, a)\Big) ^2 \Big],
\end{align}
where $\bar{\theta}_Q$ represents the parameters of the target Q-function, which is a delayed update of the Q-function, providing stable targets to reduce instability during training~\cite{mnih2013playing}.

\subsubsection{Network Architecture.}

The RL policy and the behavior policy approximator share the entire network except for the final layer. The shared network architecture follows the design of either CAtNIPP~\cite{cao2023catnipp} or the RL-based IPP~\cite{vashisth2024deep}, and any existing RL-based IPP architecture can be integrated. Here, we explain an example based on CAtNIPP. The shared network takes an augmented graph as input, where each node contains the position and the mean and variance of the GP, to generate an effective representation that captures the relationships between nodes, making the system spatially aware. This augmented graph is then processed by positional encoding. Additionally, the planning state, which includes the remaining budget and executed trajectory, is combined with the augmented graph to form a set of context-aware node embeddings. These embeddings are passed through an LSTM block, whose output is then processed by a network that employs cross-attention between the current node features and their neighboring node features. This process computes an enhanced current node representation, making it spatially aware of the environment and cognizant of the path taken to reach its current state. On top of this shared network, two separate networks are used for the approximated behavior policy and the Q-function. For the behavior policy, a final attention layer utilizes the neighboring features to extract the policy, which essentially consists of the attention scores from this layer. A binary vector mask, 
$M$, is applied to eliminate nodes that violate the budget constraint of reaching the destination from a particular node, enhancing the training process. For the Q-function, a simple MLP layer is used. Finally, the action is obtained based on Eq. \ref{eq:ourpolicy}, which requires both the approximated behavior policy and the Q-function. Our architecture is illustrated in Fig. \ref{fig:Sys_model}.  

\section{EXPERIMENTS} \label{sec:experiments}

In this section, we validate the effectiveness of our proposed approach, \texttt{OffRIPP}, in two different simulated environments: a 2D light-intensity IPP task and a 3D fruit identification task, as well as in a real-world robotic environment. Since our work is the first to apply offline RL to IPP, we generate diverse datasets with varying performance levels, comprising both optimal and sub-optimal data, using existing RL-based IPP algorithms and a traditional IPP solver for evaluation. We believe the generated dataset will also make a valuable contribution to the research community.

\begin{figure}
     \centering
     \begin{subfigure}{0.15\textwidth}
         \centering
         \includegraphics[width=\textwidth]{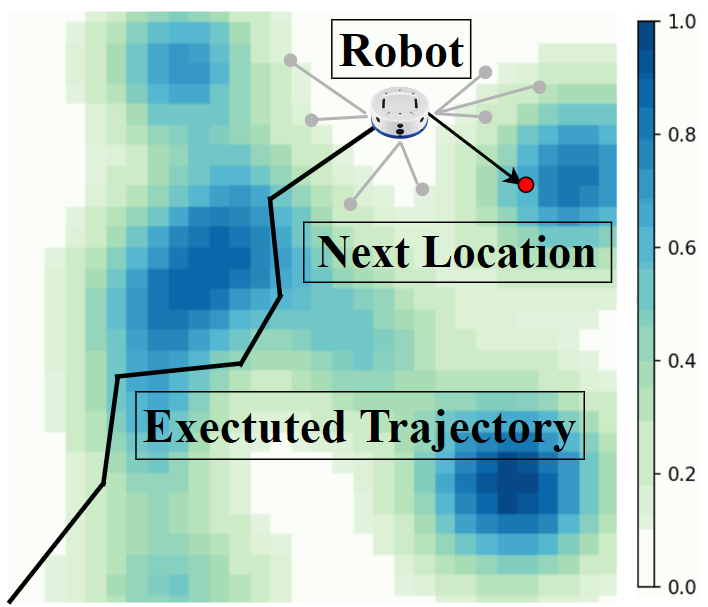}
         \caption{Light-intensity Environment}
         \label{fig:2D_env}
     \end{subfigure}
     \hfill
     \begin{subfigure}{0.15\textwidth}
         \centering
         \includegraphics[width=\textwidth]{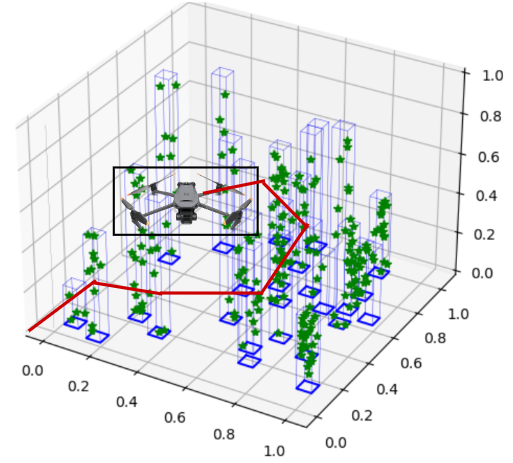}
         \caption{3D Fruit Identification Environment}
         \label{fig:3D_env}
     \end{subfigure}
     \hfill
     \begin{subfigure}{0.15\textwidth}
         \centering
         \includegraphics[width=\textwidth]{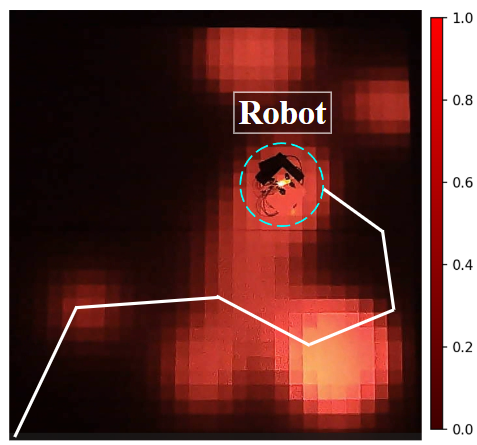}
         \caption{Real-world Experiment}
         \label{fig:real_world}
     \end{subfigure}
        \caption{Three experiments settings: (a) Heatmap representing light intensity. (b) Green stars and blue structures represent the target fruits and tree, respectively.
        (c) Real-world experiment featuring a robot (dotted circle) with the intensity map projected onto the arena.}
        \label{fig:environments}
        \vspace{-2ex}
\end{figure}

\subsection{Experiment Setup}\label{sec:env_setup}

\subsubsection{Light-intensity Environment} 
Adopted from \cite{cao2023catnipp}, the goal of this environment is to predict the light-intensity, which is the interest in the context of IPP, across the 2D map. The light intensity is constructed using $8$ to $12$ random 2-dimensional Gaussian distributions within the unit square $[0, 1]^2$. The light-intensity is illustrated in Fig. \ref{fig:2D_env}. At the start of each episode, the agent's initial prediction is initialized as a uniform distribution. During dataset collection, both the agent's start and destination positions are randomly generated within the same unit square. The environment is discretized using a PRM with a fixed number of nodes, $400$, and the number of neighboring nodes is set to $20$. The agent collects measurements every $0.2$ units of distance traveled, and while the budget is randomized between $6$ and $8$ during dataset collection, it is evaluated at fixed budgets of $6, 8,$ and $10$ during testing. The maximum episode length is fixed at 256 time steps. 

\textbf{Dataset:} We generate the following datasets using both CAtNIPP and a non-learning adaptive sampling method, with each dataset consisting of 18,500 trajectories.

$\bullet$ \textit{Expert Dataset}: This dataset is collected by deploying the CAtNIPP greedy variant's best-performing model, which has been trained for over 50,000 episodes.

$\bullet$ \textit{Medium Dataset}: This dataset is collected by deploying partially trained CAtNIPP greedy variant models, specifically those chosen between 256 and 512 episodes of training. This dataset consists of suboptimal trajectories.

$\bullet$ \textit{Greedy Planning Dataset}: This dataset is collected using a simple heuristic adaptive sampling approach, where the agent selects its next node from among its neighbors based on the highest entropy of the GP. We refer to this method as greedy planning. Note that this method is not learning-based and does not require any training.

\subsubsection{3D Fruit Identification} We use the 3D fruit identification environment introduced in \cite{vashisth2024deep}. The goal is to identify the fruits, illustrated in Fig. \ref{fig:3D_env}. The environment is represented by a $50 \times 50 \times 50$ voxel occupancy grid, which is initially unknown and updated based on sensor observations. Trees are randomly placed, and fruits are attached to the trees at random positions. The agent observes part of the field of view, including free space, observed fruits, and trees. The action space consists of four discretized yaw angles. The reward function is based on the reduction in utility uncertainty of the GP and the number of observed targets.

\textbf{Dataset:} We generate the following datasets using an RL-based IPP solver and a traditional approach, with each dataset consisting of 18,500 trajectories.

$\bullet$ \textit{Expert Dataset}: This dataset is collected by deploying a policy trained for approximately 10,000 interactions with the environment. The policy achieves the optimal performance.

$\bullet$ \textit{Medium Dataset}: This dataset is gathered using a partially trained policy, specifically models trained between 256 and 512 episodes. These policies are suboptimal and exhibit less efficient decision-making.

$\bullet$ \textit{Greedy Planning Dataset}: This dataset is collected using a simple heuristic adaptive sampling approach, similar to the 2D environment. The agent selects its next node to travel to based on the highest entropy in the GP model, without requiring any learning or training.

\begin{table}
\centering
    \centering
    \renewcommand{\arraystretch}{1.10}
    \caption{Performance comparison in terms of the trace of the covariance matrix in the light-intensity environment. A lower value indicates reduced uncertainty, indicating better performance. Note that the behavior policy is not a baseline.}
    \label{tab:2D_results}
    \begin{tabular}{cccc}
\hline
\textbf{Model} & \textbf{Budget 6} & \textbf{Budget 8} & \textbf{Budget 10} \\ \hline
\multicolumn{4}{c}{\textbf{Expert Dataset}}                      \\ \hline 
\multirow{2}{*}{\shortstack{Behavior Policy \\ (Dataset)}} & \multirow{2}{*}{18.48 ± 7.76}   & \multirow{2}{*}{6.96 ± 2.68}    & \multirow{2}{*}{3.73 ± 1.57}        \\                          &                                 &                                 &                               \\ 
\noalign{\vspace{-4pt}}
\multicolumn{4}{c}{- - - - - - - - - - - - - - - - - - - - - - - - - - - - - - - - - - - - - - - - - - - -} \\ 
\noalign{\vspace{-4pt}}
\multirow{2}{*}{\shortstack{CAtNIPP \\ (Offline Trained)}} & \multirow{2}{*}{137.5 ± 167.2} & \multirow{2}{*}{125.8 ± 203.7} & \multirow{2}{*}{110.3 ± 245.1} \\ 
                         &                                 &                                 &                               \\ 
BC                      & 30.84 ± 14.02  & 9.93 ± 5.03    & 7.02 ± 3.06        \\ 
\texttt{OffRIPP}               & \textbf{23.28} ± 5.80   & \textbf{7.83} ± 2.87    & \textbf{3.96} ± 1.41  \\ \hline \hline
\multicolumn{4}{c}{\textbf{Medium Dataset}}                      \\ \hline 
\multirow{2}{*}{\shortstack{Behavior Policy \\ (Dataset)}} & \multirow{2}{*}{32.84 ± 7.98}   & \multirow{2}{*}{18.20 ± 5.69}   & \multirow{2}{*}{9.78 ± 2.25}    \\                          &                                 &                                 &                               \\ 
\noalign{\vspace{-4pt}}
\multicolumn{4}{c}{- - - - - - - - - - - - - - - - - - - - - - - - - - - - - - - - - - - - - - - - - - - -} \\ 
\noalign{\vspace{-4pt}}
\multirow{2}{*}{\shortstack{CAtNIPP \\ (Offline Trained)}} & \multirow{2}{*}{480.8 ± 199.2} & \multirow{2}{*}{557.4 ± 207.5} & \multirow{2}{*}{628.6 ± 182.7} \\ 
                         &                                 &                                 &                               \\ 
BC                      & 48.08 ± 33.02  & 23.59 ± 19.07  & 11.36 ± 9.47    \\ 
\texttt{OffRIPP}               & \textbf{34.47} ± 27.41  & \textbf{17.91} ± 10.54  & \textbf{8.10} ± 4.42    \\ \hline \hline
\multicolumn{4}{c}{\textbf{Greedy Planning Dataset}}                      \\ \hline 
\multirow{2}{*}{\shortstack{Behavior Policy \\ (Dataset)}} & \multirow{2}{*}{73.21 ± 99.80}  & \multirow{2}{*}{65.00 ± 102.84} & \multirow{2}{*}{60.46 ± 104.41}  \\                          &                                 &                                 &                               \\ 
\noalign{\vspace{-4pt}}
\multicolumn{4}{c}{- - - - - - - - - - - - - - - - - - - - - - - - - - - - - - - - - - - - - - - - - - - -} \\ 
\noalign{\vspace{-4pt}}
\multirow{2}{*}{\shortstack{CAtNIPP \\ (Offline Trained)}} & \multirow{2}{*}{433.1 ± 169.1} & \multirow{2}{*}{517.8 ± 195.7} & \multirow{2}{*}{602.9 ± 176.5} \\ 
                         &                                 &                                 &                               \\ 
BC                      & 42.25 ± 27.91  & \textbf{16.39} ± 10.27  & 10.04 ± 5.86  \\ 
\texttt{OffRIPP}               & \textbf{39.16} ± 22.42  & 16.56 ± 12.82  & \textbf{7.61} ± 5.75  \\ \hline \hline
\multicolumn{4}{c}{\textbf{Non-learning-based IPP solvers}} \\ \hline
Greedy Planning        & 73.21 ± 99.80  & 65.00 ± 102.84 & 60.46 ± 104.41  \\ 
RAOr                   & 49.47 ± 20.29  & 19.87 ± 7.71   & 12.54 ± 5.13      \\ \hline
\end{tabular}
\end{table}

\subsection{Training Details}

To train \texttt{OffRIPP}, we use the Adam optimizer~\cite{kingma2014adam}, a batch size of 256, and an update frequency of 100 for the target network. For the threshold $\tau$ in Eq. \ref{eq:ourpolicy}
, we perform hyperparameter tuning in the range $(0, 1)$. \texttt{OffRIPP} trains for 1 epoch on the dataset, requiring an average of 3 hours in the light-intensity environment and 18 hours in the 3D fruit identification environment when using a workstation equipped with an Intel Xeon Gold 5218 CPU and three NVIDIA RTX 6000 GPUs.

\subsection{Experimental Results}\label{subsec:results}

\begin{table}
    \centering
    \renewcommand{\arraystretch}{1.10}
    \caption{Performance comparison in terms of the fruit detection rate in the 3D fruit identification environment. A higher value indicates better performance. Note that the behavior policy is not a baseline.}
    \label{tab:3D_results}
    \begin{tabular}{cccc}
    \hline
    \textbf{Model} & \textbf{Budget 6} & \textbf{Budget 8} & \textbf{Budget 10}\\ \hline
    \multicolumn{4}{c}{\textbf{Expert Dataset}}                      \\ \hline 
    \multirow{2}{*}{\shortstack{Behavior Policy \\ (Dataset)}} & \multirow{2}{*}{46.17 ± 6.94}   & \multirow{2}{*}{54.61 ± 6.41}   & \multirow{2}{*}{59.12 ± 7.82}  \\  
    &  &  &  \\ 
    \noalign{\vspace{-4pt}}
\multicolumn{4}{c}{- - - - - - - - - - - - - - - - - - - - - - - - - - - - - - - - - - - - - - - - - - - -} \\ 
\noalign{\vspace{-4pt}}
    \multirow{2}{*}{\shortstack{3D RL-based IPP \\ (Offline Trained)}} & \multirow{2}{*}{26.85 ± 9.86}   & \multirow{2}{*}{32.55 ± 10.75}  & \multirow{2}{*}{36.53 ± 12.40}   \\ 
    &  &  &  \\ 
    BC                      & 45.98 ± 7.86   & 53.87 ± 7.12   & 59.69 ± 7.51   \\ 
    \texttt{OffRIPP}               & \textbf{46.60} ± 8.87   & \textbf{55.57} ± 6.34   & \textbf{61.48} ± 6.88   \\ \hline \hline
    \multicolumn{4}{c}{\textbf{Medium Dataset}}                      \\ \hline 
    \multirow{2}{*}{\shortstack{Behavior Policy \\ (Dataset)}} & \multirow{2}{*}{27.32 ± 13.31}  & \multirow{2}{*}{32.92 ± 15.73}  & \multirow{2}{*}{38.55 ± 18.09}  \\  
    &  &  &  \\ 
    \noalign{\vspace{-4pt}}
\multicolumn{4}{c}{- - - - - - - - - - - - - - - - - - - - - - - - - - - - - - - - - - - - - - - - - - - -} \\ 
\noalign{\vspace{-4pt}}
    \multirow{2}{*}{\shortstack{3D RL-based IPP \\ (Offline Trained)}} & \multirow{2}{*}{24.31 ± 10.96}  & \multirow{2}{*}{30.45 ± 8.80}   & \multirow{2}{*}{36.21 ± 12.23}  \\  
    &  &  &  \\ 
    BC                      & 16.00 ± 6.41   & 19.86 ± 6.79   & 20.31 ± 7.82    \\ 
    \texttt{OffRIPP}               & \textbf{29.27} ± 11.08  & \textbf{32.33} ± 10.20  & \textbf{36.37} ± 11.25   \\ \hline \hline
    \multicolumn{4}{c}{\textbf{Greedy Planning Dataset}}                      \\ \hline 
    \multirow{2}{*}{\shortstack{Behavior Policy \\ (Dataset)}} & \multirow{2}{*}{17.29 ± 13.09}  & \multirow{2}{*}{17.35 ± 14.10}  & \multirow{2}{*}{31.02 ± 13.82}  \\  
    &  &  &  \\ 
    \noalign{\vspace{-4pt}}
\multicolumn{4}{c}{- - - - - - - - - - - - - - - - - - - - - - - - - - - - - - - - - - - - - - - - - - - -} \\ 
\noalign{\vspace{-4pt}}
    \multirow{2}{*}{\shortstack{3D RL-based IPP \\ (Offline Trained)}} & \multirow{2}{*}{24.40 ± 10.63}  & \multirow{2}{*}{29.43 ± 11.02}  & \multirow{2}{*}{29.41 ± 10.26}  \\  
    &  &  &  \\ 
    BC                      & 7.64 ± 7.44    & 11.39 ± 10.48  & 8.72 ± 8.34     \\ 
    \texttt{OffRIPP}               & \textbf{25.28} ± 11.31  & \textbf{29.71} ± 9.20   & \textbf{36.26} ± 10.82 \\ \hline
\end{tabular}
\end{table}

\subsubsection{Light-intensity Environment} 
For the evaluation of the proposed framework, \texttt{OffRIPP}, we consider four baselines: (i) \textit{Greedy Planning}, where the agent selects its next node from its neighbors based on the highest entropy of the GP; (ii) Randomized Anytime Orienteering (RAOr) \cite{arora2017randomized}, a sampling-based heuristic approach. Note that \textit{Greedy Planning} and RAOr do not require training, so we report their performances regardless of the dataset; (iii) Behavior Cloning (BC), which imitates the behavioral policy. We use the CAtNIPP architecture for BC and train it by optimizing the negative log-likelihood function; and (iv) CAtNIPP~\cite{cao2023catnipp} is trained offline, where the CAtNIPP agent is trained using PPO on the dataset. Besides the baselines, we also report the performance of the dataset, which reflects the quality of the data used to train the models, to validate if the trained model is able to outperform the model that generated the dataset. Note that the behavior policy is trained online, which makes comparisons with \texttt{OffRIPP} unfair.

We provide the performance of the algorithms under three types of datasets, as described in Sec. \ref{sec:env_setup}, with three different budgets of 6, 8, and 10. For the evaluation metric, we use the average trace of the covariance matrix over 50 different instances of the environment, which is the optimization objective in RL and is also used as a comparison metric in \cite{cao2023catnipp}. The corresponding results are shown in Table \ref{tab:2D_results}. It is observed that \texttt{OffRIPP} consistently produces the lowest covariance trace values across all budget instances, indicating superior performance compared to the baselines. Upon closer examination, it is noteworthy that CAtNIPP, an online RL algorithm, performs poorly, demonstrating that the naive application of RL in an offline training setting is unstable. However, \texttt{OffRIPP} performs well, even better than the dataset performance in many cases. In addition, \texttt{OffRIPP} trained on any dataset outperforms both non-learning-based algorithms, Greedy Planning and RAOr. 

\subsubsection{3D Fruit Identification Environment} For evaluation, we consider two baselines: (i) BC and (ii) the 3D RL-based IPP solver proposed in \cite{vashisth2024deep}, as detailed in Sec. \ref{subsec:ipp}. These models are assessed across three dataset types under budget constraints of 6, 8, and 10, as outlined in Sec. \ref{sec:env_setup}.

The evaluation metric used is the average fruit detection rate across 50 different instances of the environment, following the same approach for comparative analysis as employed in \cite{vashisth2024deep}. This metric captures the accuracy and efficiency of the models in detecting fruits within a limited budget, offering a reliable means of comparison. As illustrated in Table \ref{tab:3D_results}, \texttt{OffRIPP} consistently delivers superior fruit detection rates across all budget levels, significantly outperforming both BC and the 3D RL-based IPP solver. The 3D RL-based IPP solver, though designed for online RL, demonstrates suboptimal performance when applied in an offline training setting, highlighting its instability under such conditions. On the other hand, \texttt{OffRIPP} not only outperforms the baselines but also exceeds the dataset’s performance in many cases. Notably, on the Greedy Planning dataset, which is generated by a non-learning method, we observe a significant improvement.

\subsection{Analysis}

\begin{figure}
\centering
\includegraphics[width=0.9\columnwidth]{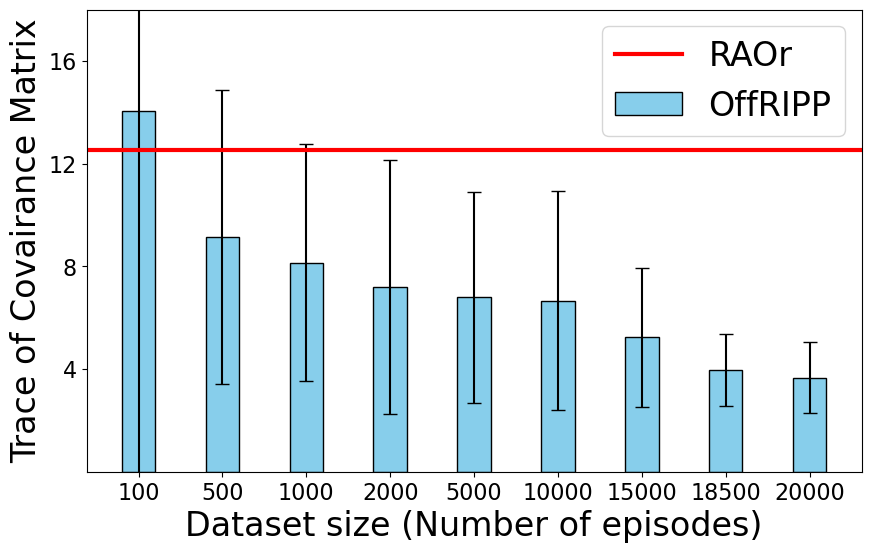}
\caption{Performance of \texttt{OffRIPP} with respect to various sizes of the dataset trained on the expert dataset and tested using a budget of 10 in the light-intensity environment. A lower value on the Y-axis is better. RAOr does not require a dataset since it is a non-learning method.}
\vspace{-1ex}
\label{fig:datasize}
\end{figure}

\subsubsection{Performance with respect to the dataset}

The performance of \texttt{OffRIPP} is influenced by both the quality and quantity of the dataset, as \texttt{OffRIPP} depends on the given dataset without generating new data based on the current policy. In Table~\ref{tab:2D_results} and Table~\ref{tab:3D_results}, it is observed that performance improves as the performance of the behavior policy improves. We now provide the performance with respect to the dataset size. Fig.~\ref{fig:datasize} shows the performance of \texttt{OffRIPP} and RAOr. Performance improves as the number of training episodes increases, with convergence beginning around $18k$ episodes. This suggests that $18k$ is the minimum dataset size required for optimal performance. Furthermore, \texttt{OffRIPP} requires only $500$ episodes to outperform RAOr, a non-learning-based IPP solver. 

\subsubsection{Total Planning time}

One of the advantages of \texttt{OffRIPP} is its fast planning during the execution phase. We provide the total planning time for executing one trajectory for \texttt{OffRIPP}, BC, and RAOr evaluated in the light-intensity environment. \texttt{OffRIPP} requires 0.64 seconds, slightly more time compared to BC (0.62 seconds) due to the additional network—estimation of Q-value—however, it is significantly faster than RAOr (6.11 seconds), the SOTA non-learning-based IPP solver. Therefore, \texttt{OffRIPP} outperforms the non-learning method in both performance and planning time.

\subsection{Experimental Validation on Real Robot}

We validate \texttt{OffRIPP} on a real-robot system in a light-intensity environment. Specifically, we project the light-intensity environment onto a physical $1.5 \times 1.5 \  \text{m}^2$ arena, as shown in Fig. \ref{fig:real_world}, maintaining consistency in experimental configuration between simulation and physical deployment. The Khepera-IV robot, equipped with a Raspberry Pi 3 and a camera module, is then deployed. In this experiment, we first train the robot using \texttt{OffRIPP} with the expert dataset, and then deploy it in the arena. During deployment, the robot collects real-time images and executes actions from those observations. This experiment demonstrates the adaptability of \texttt{OffRIPP}, enabling rapid decision-making and action in real time. Our results confirm \texttt{OffRIPP}'s suitability for real-world use, showing it can generalize from simulation to reality without retraining or extensive modifications. A video of this experiment is provided\footnote{The link for the real-robot experiment - \href{https://drive.google.com/file/d/10DyngZOL6V8Z46AwYgk6LI7EHs-9Fhdx/view?usp=sharing}{\color{blue}{Link}.}}.

\section{CONCLUSION} \label{sec:conclusion}

In this paper, we present an offline RL-based IPP framework that enables training an RL agent to find a path that maximizes information gain without environment interactions during training. Based on the IPP architecture, which incorporates a graph, environment modeling using a GP, and attention mechanisms, we model an approximated behavior policy that imitates the policy used to generate the dataset, along with a Q-function. We then build a policy that chooses an action that maximizes the Q value, using the approximated behavior policy to select from among the actions likely found in the dataset. We show that \texttt{OffRIPP} outperforms the baselines, including the traditional IPP solver, in terms of both performance and planning time across two different environments. Additionally, we validate the effectiveness of \texttt{OffRIPP} in a real-robot system. Future work includes learning multi-agent IPP policies from a dataset. 

\section*{ACKNOWLEDGMENT}

This work has been supported by NSF and USDA-NIFA under AI Institute for Resilient Agriculture, Award No. 2021-67021-35329 and Department of Agriculture Award Number 2023-67021-39073. The authors would like to thank the RISS program for providing the opportunity to conduct this research. A special thanks to Mrs. Rachel Burcin, Dr. John M. Dolan, Ms. Morgan Grimm, and the entire CMU Robotics Institute community for their support. The authors also gratefully acknowledge Dr. Simon Stepputtis for his assistance in conducting real-world experiments.

\bibliographystyle{IEEEtran}
\bibliography{ref}

\end{document}